\pdfoutput=1
\documentclass[11pt]{article}

\usepackage[final]{acl}

\usepackage{times}
\usepackage{latexsym}
\usepackage{amsmath}
\usepackage{amssymb}
\usepackage{bm}

\usepackage[T1]{fontenc}
\usepackage[utf8]{inputenc}

\usepackage{microtype}

\usepackage{inconsolata}
\usepackage{tcolorbox}
\usepackage{booktabs}

\usepackage{graphicx}
\usepackage{subcaption}
\usepackage{caption}

\title{Boosting CTC-Based ASR Using LLM-Based Intermediate Loss Regularization}

\author{Duygu Altinok \\
  Independent Researcher, Germany \\
  \texttt{duygu.altinok@onlyduygu.com}}

\begin{document}
\maketitle
\begin{abstract}
End-to-end (E2E) automatic speech recognition (ASR) systems have revolutionized the field by integrating all components into a single neural network, with attention-based encoder-decoder models achieving state-of-the-art performance. However, their autoregressive decoding process limits inference speed, making them unsuitable for real-time applications. In contrast, CTC-based models offer faster, non-autoregressive decoding but struggle to model linguistic dependencies effectively. Addressing this challenge, we propose a novel auxiliary loss framework called Language-Aware Intermediate Loss (LAIL) to enhance CTC-based ASR using the linguistic knowledge of large language models (LLMs). By attaching connector layers to intermediate encoder layers, LAIL maps outputs to the embedding space of an LLM and computes a causal language modeling loss during training. This approach enhances linguistic modeling while preserving the computational efficiency of CTC decoding. Using the Conformer architecture and various LLaMA models, we demonstrate significant improvements in Word Error Rate (WER) on the LibriSpeech, TEDLIUM2, and WSJ corpora, achieving state-of-the-art performance for CTC-based ASR with minimal computational overhead.
\end{abstract}
\section{Introduction}
The field of automatic speech recognition (ASR) has seen significant advancements with the introduction of end-to-end (E2E) ASR systems powered by deep neural networks \cite{allyouneed,8462105,8462506,7472618}. Unlike traditional hybrid ASR architectures, which rely on separate components such as acoustic, lexicon, and language models, E2E systems simplify the pipeline by integrating all components into a single neural network. This architectural simplicity, combined with high performance, has made E2E ASR models a popular choice in modern ASR research and applications.

Among E2E ASR architectures, the attention-based encoder-decoder model has emerged as a dominant paradigm \cite{pmlr-v32-graves14,9053889}. These models generate output tokens sequentially in an autoregressive (AR) manner, exploiting contextual dependencies between tokens. As a result, attention-based models achieve state-of-the-art recognition performance on multiple benchmarks \cite{radford2022robustspeechrecognitionlargescale,gulati2020conformer,squeeze,fastconfo}. However, their autoregressive decoding process leads to slower inference speeds, making them less suitable for real-time or resource-constrained applications.

On the other hand, Connectionist Temporal Classification (CTC)-based models \cite{ctc-graves,karita2019improving} offer a simpler and more efficient alternative for E2E ASR. These models align input speech sequences to output label sequences using a probabilistic approach that eliminates the need for explicit frame-level annotations. Moreover, CTC-based models perform inference in a non-autoregressive (NAR) manner, enabling faster decoding compared to attention-based models. Despite these advantages, their performance often lags behind attention-based models due to the conditional independence assumption between output tokens, which limits their ability to effectively model linguistic dependencies.

Improving the recognition performance of CTC-based models without sacrificing their computational efficiency is a critical area of research. Various approaches have been proposed to address this challenge \cite{nartrans,higuchi2020mask,chi-etal-2021-align,ctc-gated,hierarchical-ctc}. Notable among these are Intermediate CTC \cite{ctc-inter}, which introduces auxiliary CTC losses at intermediate encoder layers, and Self-Conditioned CTC \cite{nozaki2021relaxing}, which conditions the final predictions on intermediate outputs to partially relax the conditional independence assumption. These methods achieve notable improvements while retaining fast inference by modifying only the encoder, without introducing iterative decoding mechanisms.

Motivated by the success of auxiliary loss techniques, we propose leveraging the vast linguistic knowledge of Large Language Models (LLMs) to further enhance the performance of CTC-based ASR. LLMs, pre-trained on massive text corpora, excel in language-related tasks such as syntax understanding, semantic reasoning, and vocabulary modeling \cite{openai2024gpt4technicalreport,touvron2023llamaopenefficientfoundation}. Integrating such knowledge into CTC-based models can help bridge the gap between their speed and the linguistic capabilities of attention-based models.

Building on the auxiliary loss framework introduced by \cite{ctc-inter}, we aim to incorporate the rich textual knowledge of LLMs into CTC-based ASR. Specifically, we propose an Language-Aware Intermediate Loss (LAIL), where we attach connector layers at selected encoder layers of a CTC-based model to map their outputs to the embedding space of an LLM. The CLM Loss is computed as a causal language model loss \cite{brown2020languagemodelsfewshotlearners} between the ground truth transcript and the LLM’s outputs, where each token prediction depends only on previous tokens.  This enables the computation of a causal language modeling loss between the ground truth transcript and the LLM's outputs, effectively embedding linguistic knowledge into the ASR model during training.

Our approach retains the computational efficiency of CTC-based inference by keeping the decoding process unchanged. At inference, the model uses standard CTC decoding with greedy search, avoiding the autoregressive bottleneck of LLMs while benefiting from their linguistic knowledge during training. We employ the Conformer architecture as the encoder and experiment with various configurations of LLMs from the LLaMA family (1B, 3B, and 8B parameters). We also explore the impact of attaching auxiliary losses to different encoder layers (e.g., bottom, top, or intermediate layers) and varying the number of selected layers.

Our key contributions are summarized as follows:
\begin{itemize}
    \item We propose a simple yet effective auxiliary loss, called Language-Aware Intermediate Loss (LAIL), to enhance the performance of CTC-based ASR models.
    \item We demonstrate the effectiveness of integrating the linguistic knowledge of LLMs (LLaMA models) into the Conformer-CTC architecture.
    \item We investigate the impact of attaching auxiliary losses to different encoder layers and show that a few layers close to the final encoder layers are sufficient for significant performance gains.
    \item Our approach maintains the computational efficiency of CTC-based decoding, ensuring fast inference without autoregressive bottlenecks.
\end{itemize}

We evaluate our method on the LibriSpeech \cite{7178964}, TEDLIUM2 \cite{rousseau-etal-2012-ted} and Wall Street Journal (WSJ) \cite{paul-baker-1992-design} corpora, demonstrating significant improvements in WER compared to baseline models. Our results highlight the potential of integrating LLM-based intermediate losses to enhance CTC-based ASR while preserving its decoding speed. By leveraging connector layers with auxiliary losses, we effectively combine the strengths of LLMs and CTC models, bridging the gap between linguistic knowledge and computational efficiency.

\begin{figure*}[ht]
    \centering
    \includegraphics[width=\textwidth]{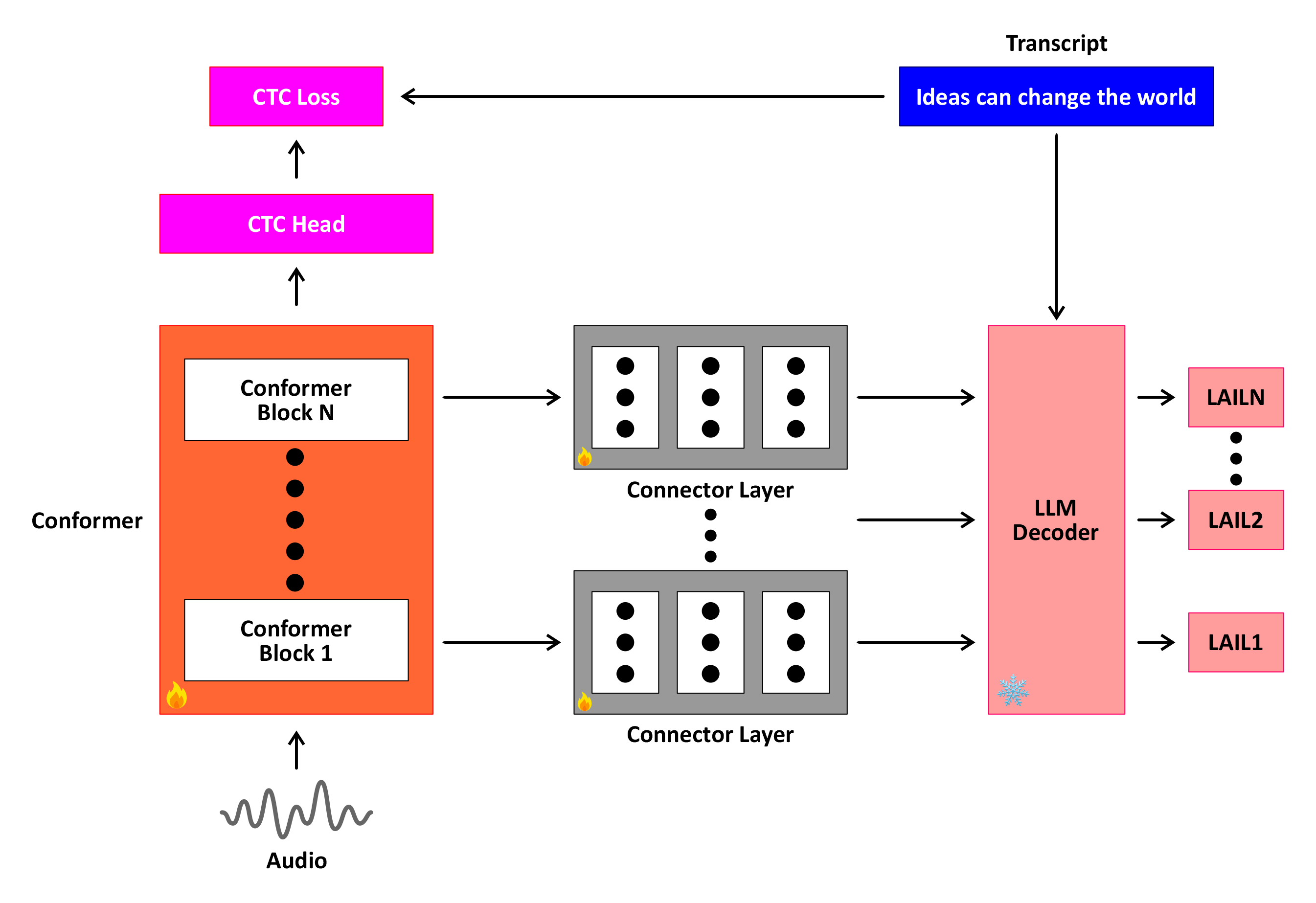} 
    \caption{Overall architecture of the proposed model. Auxiliary losses (LAIL) are attached after selected Conformer blocks. The Conformer model includes a CTC head at the top, and training is performed jointly, with the final loss computed as a weighted sum of the CTC loss and LAILs.}
    \label{fig:spread-entities} 
\end{figure*}

\section{Background and Related Work}
\subsection{CTC}

CTC calculates the probability of the target sequence $y$ by considering all possible alignments between the labels and the input sequence of length \( T \). Given the encoder output \( \mathbf{x}_L \) and the target sequence \( y \), the likelihood is defined as:

\begin{equation}
P_{\text{CTC}}(y|\mathbf{x}_L) = \sum_{a \in \beta^{-1}(y)} P(a|\mathbf{x}_L)
\end{equation}

where \( \beta^{-1}(y) \) represents the set of alignments \( a \) of length \( T \) that are compatible with \( y \), including the special blank token. The alignment probability \( P(a|\mathbf{x}_L) \) is factorized using the following conditional independence assumption:

\begin{equation}
P(a|\mathbf{x}_L) = \prod_{t} P(a[t]|\mathbf{x}_L[t])
\end{equation}

Here, \( a[t] \) and \( \mathbf{x}_L[t] \) denote the \( t \)-th symbol of the alignment \( a \) and the \( t \)-th representation vector of \( \mathbf{x}_L \), respectively.

During training, the negative log-likelihood computed using \( P_{\text{CTC}}(y|\mathbf{x}_L) \) in (1) is minimized as follows:

\begin{equation}
\mathcal{L}_{\text{CTC}} = -\log P_{\text{CTC}}(y|\mathbf{x}_L)
\end{equation}

During inference, decoding can be performed using various strategies, including greedy search for fast and efficient alignment, prefix search combined with a language model (LM) to incorporate linguistic constraints, or beam search with an LM to explore multiple hypotheses simultaneously and achieve a better balance between accuracy and computational complexity. Additionally, other advanced variants, such as shallow fusion or rescoring with an external LM, can be employed to further refine the decoding process.

\subsection{NAR and AR}
Given a source acoustic feature sequence \( \mathbf{X} = (x_1, \cdots, x_{T'}) \), an ASR model predicts the target transcription sequence \( \mathbf{Y} = (y_1, \cdots, y_T) \) using the conditional distribution:

\begin{equation}
P(\mathbf{Y}|\mathbf{X}; \theta) = \prod_{t=1}^{T} P_{\text{AR}}(y_t|y_{<t}, \mathbf{X}; \theta)
\end{equation}

where \( \theta \) represents the model parameters. This autoregressive (AR) generation approach captures sequential dependencies in the transcription but suffers from high inference latency due to its token-by-token prediction process.

In contrast, non-autoregressive (NAR) models perform conditionally independent predictions for parallel transcription \cite{comparative}:

\begin{equation}
P(\mathbf{Y}|\mathbf{X}; \theta) = \prod_{t=1}^{T} P_{\text{NAR}}(y_t|\mathbf{X}; \theta)
\end{equation}

While the vanilla NAR models accelerates transcription like 15x, 30x and 40x \cite{comparative}, it still shows a significant performance gap compared to its AR counterpart.

\subsection{Conformer}
In this work, we utilized the Conformer \cite{gulati2020conformer} model as the acoustic encoder. The Conformer is an advanced variant of the Transformer \cite{allyouneed} architecture. The Transformer leverages self-attention \cite{allyouneed} (denoted as \( \text{SelfAttention}(\cdot) \) in Eq. (\ref{eq:transformer_self_attention})) to capture global representations, while layer normalization \cite{ba2016layer} and residual connections \cite{deepres} are incorporated to stabilize the training process. In the Transformer architecture, the encoder layer \( \text{EncoderLayer}(\cdot) \) in Eq. (\ref{eq:encoder_layer_transformer}) is defined as:

\begin{equation}
\mathbf{x}_\text{MHA}^{(l)} = \text{SelfAttention}(\mathbf{x}^{(l-1)}) + \mathbf{x}^{(l-1)}, 
\label{eq:transformer_self_attention}
\end{equation}

\begin{equation}
\mathbf{x}^{(l)} = \text{FFN}(\mathbf{x}_\text{MHA}^{(l)}) + \mathbf{x}_\text{MHA}^{(l)},
\label{eq:encoder_layer_transformer}
\end{equation}

where \( \text{FFN}(\cdot) \) represents the feed-forward network layers.

The Conformer extends the Transformer by integrating convolutional neural network (CNN) layers, enabling efficient learning of both global and local representations. In the Conformer architecture, the encoder layer \( \text{EncoderLayer}(\cdot) \) in Eq. (\ref{eq:encoder_layer_conformer}) is defined as:

\begin{equation}
\mathbf{x}_\text{FFN}^{(l)} = \frac{1}{2} \text{FFN}(\mathbf{x}^{(l-1)}) + \mathbf{x}^{(l-1)}, 
\label{eq:conformer_ffn_1}
\end{equation}

\begin{equation}
\mathbf{x}_\text{MHA}^{(l)} = \text{SelfAttention}(\mathbf{x}_\text{FFN}^{(l)}) + \mathbf{x}_\text{FFN}^{(l)},
\label{eq:conformer_self_attention}
\end{equation}

\begin{equation}
\mathbf{x}_\text{Conv}^{(l)} = \text{Convolution}(\mathbf{x}_\text{MHA}^{(l)}) + \mathbf{x}_\text{MHA}^{(l)}, 
\label{eq:conformer_convolution}
\end{equation}

\begin{equation}
\mathbf{x}^{(l)} = \text{LayerNorm}\left(\frac{1}{2} \text{FFN}(\mathbf{x}_\text{Conv}^{(l)}) + \mathbf{x}_\text{Conv}^{(l)}\right).
\label{eq:encoder_layer_conformer}
\end{equation}

Here, \( \text{Convolution}(\cdot) \) denotes the convolutional layers, and \( \text{LayerNorm}(\cdot) \) represents the layer normalization operation. The Conformer architecture effectively combines self-attention and convolution to model both contextual and local dependencies in the input sequence.

\subsection{LLMs and CTC-based ASR}
The integration of large language models (LLMs) with CTC-based ASR has been relatively underexplored. However, earlier works have attempted to incorporate pre-trained language models, such as BERT \cite{devlin-etal-2019-bert}, into CTC frameworks. For instance, \cite{futami2022distillingknowledgebertctcbased} distilled BERT's linguistic knowledge into CTC-based ASR by extending knowledge distillation to frame-synchronous CTC models. This approach enabled the integration of BERT's contextual representations while retaining the fast inference characteristic of CTC. In their method, BERT served as a teacher model, predicting masked words based on both left and right contexts.

Another notable study, \cite{higuchi2020mask}, also leveraged BERT to enhance CTC models by relaxing the conditional independence assumption inherent in traditional CTC. This work introduced BERT-CTC, which incorporates linguistic knowledge through BERT's contextual embeddings. BERT-CTC attends to the full context of input and output sequences using a self-attention mechanism, encouraging the model to capture interdependencies between audio and token representations while maintaining CTC's training efficiency. During inference, BERT-CTC employs a mask-predict algorithm combined with CTC decoding to iteratively refine the output sequence.

More recently, \cite{ctc-llm} utilized the power of LLMs for contextuality in ASR. They proposed a filtering algorithm that uses coarse decoding results from a fine-tuned self-supervised speech model with a simple CTC head to identify and select relevant hotwords. These filtered hotwords were then incorporated into LLM prompts, resulting in improved recognition performance. This method demonstrated the potential of combining LLMs with CTC for enhanced contextual understanding and significant performance gains.

\section{Proposed Method}
This section outlines the details of the architectural components and the methodology used in our approach.
\subsection{Model}
\subsubsection{LLM}

We employed a pretrained LLaMA 3 model with 8 billion parameters as the large language model (LLM) in our experiments. The LLaMA model was chosen for its exceptional performance in language-related tasks, including syntax understanding and semantic reasoning.
\subsubsection{Conformer-CTC}

For the ASR component, we utilized the Wav2Vec 2.0 Conformer Large model \cite{baevski2020wav2vec}, comprising 370 million parameters \footnote{\url{https://huggingface.co/facebook/wav2vec2-conformer-rope-large-960h-ft}}. This model features rotary positional embeddings, a hidden dimensionality of 1024, a kernel size of 9, and a stacked architecture of 24 encoder blocks. It is pretrained using a contrastive loss function and further fine-tuned on 960 hours of LibriSpeech data with a CTC loss to optimize it for ASR tasks.
\subsubsection{Connector Layers}

The connector layers serve as the bridge between the Conformer encoder and the LLaMA model. These layers start with five down-sampling blocks, which reduce the temporal resolution of the audio representations by a factor of 32. The final stage is a linear layer that maps the output to the LLaMA model’s embedding space. As a result, the decoder processes audio tokens sampled every 320ms, with each token represented as a 4096-dimensional vector, aligning with the embedding size of the LLaMA model.

\subsection{Language-Aware Intermediate Loss (LAIL)}

To incorporate the linguistic knowledge of the LLM into the Conformer-CTC model, we propose a novel auxiliary loss called Language-Aware Intermediate Loss (LAIL). This loss is computed at selected intermediate layers of the Conformer encoder to align its representations with the embedding space of the LLM. The LAIL encourages the Conformer encoder to learn linguistically informed representations during training, thereby enhancing its transcription capabilities.

Given the output of an intermediate encoder layer of the Conformer, denoted as \( \mathbf{h}_l \in \mathbb{R}^{T \times d} \), where \( T \) is the sequence length and \( d \) is the dimensionality of the hidden representation, the projection layer maps \( \mathbf{h}_l \) to the LLM embedding space:

\begin{equation}
\mathbf{z}_l = \text{Linear}(\mathbf{h}_l) \in \mathbb{R}^{T \times d_{\text{LLM}}},
\end{equation}

where \( d_{\text{LLM}} \) is the embedding dimensionality of the LLM.

The LLM computes token probabilities using a causal language modeling objective. Let the ground truth transcript tokens be \( \mathbf{y} = (y_1, y_2, \dots, y_{N}) \), where \( N \) is the length of the transcript. The causal language modeling loss is defined as:

\begin{equation}
\mathcal{L}_{\text{CLM}} = - \sum_{t=1}^{N} \log P_{\text{LLM}}(y_t | y_{<t}, \mathbf{z}_l),
\end{equation}

where \( P_{\text{LLM}}(y_t | y_{<t}, \mathbf{z}_l) \) represents the probability of token \( y_t \), conditioned on all preceding tokens \( y_{<t} \) and the LLM-aligned embeddings \( \mathbf{z}_l \).

The total Language-Aware Intermediate Loss is computed as a weighted sum over the selected intermediate layers \( L \) of the Conformer encoder:

\begin{equation}
\label{eqn:final-loss}
\mathcal{L}_{\text{LAIL}} = \sum_{l \in L} \lambda_l \mathcal{L}_{\text{CLM}, l},
\end{equation}

where \( \lambda_l \) is the weight assigned to the auxiliary loss at layer \( l \), and \( \mathcal{L}_{\text{CLM}, l} \) is the CLM loss computed at layer \( l \).

Finally, the overall training objective combines the CTC loss and the LAIL:

\begin{equation}
\mathcal{L}_{\text{total}} = \mathcal{L}_{\text{CTC}} + \alpha \mathcal{L}_{\text{LAIL}},
\end{equation}

where \( \alpha \) is a hyperparameter that controls the contribution of the LAIL to the total loss. This formulation allows the Conformer-CTC model to benefit from both its inherent alignment capabilities and the linguistic knowledge embedded in the LLM.

\subsection{Datasets} We experimented on LibriSpeech, TEDLIUM2 and WSJ corpora. These datasets include 1000, 207 and 81 hours of English audio respectively.

\section{Experimental Setup}

Our experiments were conducted in two main steps:
\begin{itemize}
    \item \textbf{Fine-tuning the Conformer model} on the target datasets (excluding LibriSpeech) using the CTC loss.
    \item \textbf{Training the architecture} with the Conformer and connector layers, while keeping the LLaMa layers frozen.
\end{itemize}

\subsubsection{Fine-Tuning the Conformer}
We fine-tuned the Conformer model using the HuggingFace Trainer \cite{wolf2020huggingfacestransformersstateoftheartnatural} with a batch size of $32$ and a learning rate of $2e-5$. To prevent overfitting, we limited the number of epochs to $10$ for WSJ (a smaller dataset) and $15$ for TEDLIUM 2 (a larger dataset). For both datasets, we applied $1250$ warm-up steps to stabilize early training. We will call these models as Conformer-tuned.

Fine-tuning was performed using the AdamW optimizer with a weight decay of $0.01$. To mitigate catastrophic forgetting, the latent feature extractor was frozen during the first 20\% of the epochs, after which it was unfrozen and trained for the remaining epochs.

\subsubsection{Placing the Connector Layers}  
We experimented with different configurations for placing the connector layers by attaching varying numbers of loss heads at different encoder positions. Specifically, we tested setups with \textbf{5, 4, 3, 2, and 1 loss head(s)}, as detailed below:

\begin{itemize}
    \item \textbf{5 heads}: Placed after Conformer blocks 4, 8, 16, 20, and 24.
    \item \textbf{4 heads}: Placed after Conformer blocks 6, 12, 18, and 24.
    \item \textbf{3 heads}: Placed after Conformer blocks 8, 16, and 24.
    \item \textbf{2 heads}: Experimented with placements targeting different encoder regions:
    \begin{itemize}
        \item Bottom and top layers: Blocks 6 and 24.
        \item Middle and top layers: Blocks 12 and 24.
        \item Upper layers: Blocks 18 and 24.
    \end{itemize}
    \item \textbf{1 head}: Placed after the 24th (final) Conformer block.
\end{itemize}

These configurations allowed us to evaluate the impact of connector layer placement on model performance and understand the contribution of intermediate and final encoder layers.

\subsubsection{Training the Full Architecture}
In the second step, we trained the connector layers and the Conformer encoder jointly, while keeping the LLaMa layers frozen. The batch size was set to $32$, and the learning rate was $5e-5$. To prevent overfitting, we trained for $3$ epochs per dataset. We will call these models Conformer-LAIL.

\subsubsection{Hardware and Code Availability}
All experiments were conducted on a single NVIDIA H100 GPU. Our implementation is publicly available on GitHub\footnote{\url{https://github.com/DuyguA/TSD2025-CTC-LLM-based-Loss-Regularization}}.

\section{Results and Discussion}
Table \ref{tab:base-stats} presents the WER results on various datasets, comparing the baseline fine-tuned Conformer model (Conformer-tuned) with our proposed Conformer-LAIL model using 4 connector heads and a loss weight of $\alpha = 0.3$ as defined in \ref{eqn:final-loss}. The value of $\alpha$ was determined through a hyperparameter search over the range $[0.0, 1.0]$, and all results were obtained using the LLaMa 3 model with 8B parameters.

\setlength{\tabcolsep}{2pt}
\begin{table}[th]
\begin{tabular}{l|c|c}
\toprule
\textbf{Dataset}       & \textbf{Conformer-tuned}  & \textbf{Conformer-LAIL}    \\
\midrule
LibriSpeech             &                          &                           \\
\hspace{5mm}test-clean  & 1.96                     & 1.74                      \\
\hspace{5mm}test-other  & 3.98                     & 2.96                      \\
\midrule
TEDLIUM 2              & 7.7                      & 6.0                       \\
WSJ                     & 5.1                      & 3.6                       \\
\bottomrule
\end{tabular}
\caption{WER results on various datasets achieved by the fine-tuned Conformer vs. Conformer-LAIL models.}
\label{tab:base-stats}
\end{table}

LibriSpeech is a large, well-balanced dataset with approximately 960 hours of clean and high-quality audio. It is considered the easiest dataset among the three. On the \texttt{test-clean} subset, Conformer-LAIL improves the baseline by over 10\% absolute. For the more challenging \texttt{test-other} subset, which contains noisier and more complex audio, our method achieves a significant improvement of 25\%. This is expected, as the noisier subset benefits more from the linguistic knowledge provided by LLaMA.

TEDLIUM2 is a more diverse and challenging dataset compared to LibriSpeech, featuring spontaneous speech and accents from TED talks. Despite these challenges, Conformer-LAIL achieves a 22\% improvement over the baseline, demonstrating its ability to generalize well to diverse linguistic and acoustic content.

WSJ contains clean, formal speech with domain-specific vocabulary focused on financial and business topics. While it is acoustically simpler, its smaller size and specialized vocabulary present additional challenges. Here, Conformer-LAIL achieves the largest relative improvement of 29\%, likely due to LLaMA's extensive vocabulary and textual knowledge, which includes domain-specific content such as financial terminology.

\subsection{Effect of Number and Placement of Connector Layers}
We investigated the effect of the number and placement of connector layers by experimenting with various configurations. Starting with 4 connectors placed at layers 6, 12, 18, and 24, we hypothesized that this configuration would capture multi-granular features and optimize linguistic alignment.  

\begin{itemize}
    \item \textbf{Intermediate Representations:} Lower layers (e.g., layer 6) capture acoustic features like phonemes, while mid-to-high layers (e.g., 12, 18) encode abstract features such as word- and phrase-level semantics. The final layer (24) provides fully contextualized, sentence-level representations. This progression ensures multi-granular alignment between Conformer outputs and LLaMA embeddings.  
    \item \textbf{Linguistic Alignment:} Using connectors across progressively deeper layers (6, 12, 18, and 24) improves alignment of acoustic features with linguistic structures. The final layer ensures contextually complete representations for optimizing LAIL.  
    \item \textbf{Balanced Computational Efficiency:} Placing connectors on a subset of layers (rather than all layers) minimizes computational overhead while maintaining performance.  
\end{itemize}  

Table \ref{tab:head-results} shows the WER results for different connector configurations.  

\begin{table*}[th]
\centering
\begin{tabular}{l|c|c|c|c}
\toprule
\textbf{Connector layers} & \textbf{LS test-clean} & \textbf{LS test-other}  & \textbf{TEDLIUM2}  & \textbf{WSJ}  \\
4,8,12,16,20,24 & 1.70 & 2.90 & 5.95 & 3.2 \\ 
\midrule
6,12,18,24 & 1.74 & 2.96 & 6.00 & 3.60 \\
\midrule
8,16,24 &  1.76 &  3.00 & 6.20  & 3.70 \\
\midrule
6,24    & 1.80  & 3.10  & 6.45  & 4.24 \\ 
\midrule
12,24   & 1.85  & 3.21  & 6.67  & 4.00  \\
\midrule 
18,24    & 1.86  & 3.43  & 6.78 &  3.80 \\
\midrule
24    &  1.90   & 3.55 & 6.86 &  4.32 \\
\bottomrule
\end{tabular}
\caption{Effect of connector layer placement on WER across datasets. LS denotes LibriSpeech.}
\label{tab:head-results}
\end{table*}

The results indicate that the number and placement of connectors significantly impact WER:  
\begin{itemize}
    \item \textbf{5 heads (4,8,12,16,20,24):} Provides the most granular alignment but increases computational overhead.
    \item \textbf{4 heads (6,12,18,24):} Balances granularity and efficiency, yielding consistent improvements across datasets.  
    \item \textbf{3 heads (8,16,24)}: Focuses on progressively deeper features, slightly sacrificing performance for reduced computation.  
    \item \textbf{2 heads (6,24 or 12,24):} Reduces alignment granularity; however, configurations like 6,24 capture both early acoustic and high-level linguistic features.  
    \item \textbf{1 head (24):} Focuses solely on final linguistic representations, missing earlier acoustic and mid-level cues, resulting in the highest WER.  
\end{itemize}

Interestingly, WSJ exhibited a different trend compared to other datasets. Higher-layer connectors (e.g., 18,24 or 24 alone) performed better, while lower-layer replacements degraded performance. This aligns with WSJ's domain-specific vocabulary and structure, which benefit more from LLM embeddings at higher layers. Removing high-layer connectors hurt WSJ performance significantly more than removing lower-layer connectors, underscoring the importance of LLM-driven linguistic knowledge for this dataset.

\subsection{Effect of LLM Size on Results}  
To evaluate the impact of LLM size, we compared the performance of Conformer-LAIL when using LLaMa models with 1B, 3B, and 8B parameters. For these experiments, the number of connector layers was fixed at 4 (placed at layers 6, 12, 18, and 24), and the loss weight \(\alpha\) was set to 0.3. The results are summarized in Table \ref{tab:llm-size-results}.  

\begin{table*}[th]
\centering
\begin{tabular}{l|c|c|c|c}
\toprule
\textbf{LLM Size} & \textbf{LS test-clean} & \textbf{LS test-other} & \textbf{TEDLIUM2} & \textbf{WSJ} \\
\midrule
1B                 & 1.82    & 3.37       & 6.28              & 4.50         \\
3B                 & 1.78    & 3.10          & 6.12              & 3.90         \\
8B                 & 1.74    & 2.96       & 6.00              & 3.60         \\
\bottomrule
\end{tabular}
\caption{Effect of LLM size on WER for Conformer-LAIL across datasets.}
\label{tab:llm-size-results}
\end{table*}  

The results show a clear trend: as the size of the LLM increases, the WER decreases across all datasets.  

\begin{itemize}
 \item \textbf{1B parameters:} While the smallest LLM still improves over the baseline Conformer model, the gains are modest, particularly for more linguistically challenging datasets such as WSJ.  
 \item \textbf{3B parameters:} This intermediate-sized LLM provides a notable improvement by better capturing linguistic structures and domain-specific knowledge, especially for WSJ and TEDLIUM2.   
 \item \textbf{8B parameters:} The largest model achieves the best performance across all datasets, leveraging its extensive vocabulary and linguistic reasoning capacity to enhance alignment with Conformer outputs. 
\end{itemize}
   
These findings highlight that larger LLMs provide richer linguistic embeddings, which significantly benefit datasets with diverse or domain-specific linguistic structures (e.g., WSJ). However, the trade-off is increased computational cost, which must be considered in resource-constrained environments.

\section{Conclusion}  
In this work, we proposed Language-Aware Intermediate Loss (LAIL), a novel auxiliary loss framework that integrates the linguistic knowledge of LLMs into CTC-based ASR systems. By attaching connector layers at selected Conformer encoder blocks and leveraging the embeddings of LLaMa models, our approach enhances linguistic alignment while maintaining the computational efficiency of CTC-based inference. Extensive experiments on LibriSpeech, TEDLIUM2, and WSJ datasets demonstrate significant WER improvements, with larger LLMs (e.g., LLaMa 8B) providing the best performance. Our results highlight the potential of combining LLMs with CTC to bridge the gap between linguistic richness and computational efficiency, offering a promising direction for future ASR research.

\begin{raggedright}
\bibliography{custom}

\begin{thebibliography}{33}
\providecommand{\natexlab}[1]{#1}

\bibitem[{Ba et~al.(2016)Ba, Kiros, and Hinton}]{ba2016layer}
Jimmy~Lei Ba, Jamie~Ryan Kiros, and Geoffrey~E. Hinton. 2016.
\newblock \href {http://arxiv.org/abs/1607.06450} {Layer normalization}.
\newblock Cite arxiv:1607.06450.

\bibitem[{Baevski et~al.(2020)Baevski, Zhou, Mohamed, and Auli}]{baevski2020wav2vec}
Alexei Baevski, Yuhao Zhou, Abdelrahman Mohamed, and Michael Auli. 2020.
\newblock wav2vec 2.0: A framework for self-supervised learning of speech representations.
\newblock In \emph{Advances in Neural Information Processing Systems}, volume~33, pages 12449--12460.

\bibitem[{Bahdanau et~al.(2016)Bahdanau, Chorowski, Serdyuk, Brakel, and Bengio}]{7472618}
Dzmitry Bahdanau, Jan Chorowski, Dmitriy Serdyuk, Philémon Brakel, and Yoshua Bengio. 2016.
\newblock \href {https://doi.org/10.1109/ICASSP.2016.7472618} {End-to-end attention-based large vocabulary speech recognition}.
\newblock In \emph{2016 IEEE International Conference on Acoustics, Speech and Signal Processing (ICASSP)}, pages 4945--4949.

\bibitem[{Brown et~al.(2020)Brown, Mann, Ryder, Subbiah, Kaplan, Dhariwal, Neelakantan, Shyam, Sastry, Askell, Agarwal, Herbert-Voss, Krueger, Henighan, Child, Ramesh, Ziegler, Wu, Winter, Hesse, Chen, Sigler, Litwin, Gray, Chess, Clark, Berner, McCandlish, Radford, Sutskever, and Amodei}]{brown2020languagemodelsfewshotlearners}
Tom~B. Brown, Benjamin Mann, Nick Ryder, Melanie Subbiah, Jared Kaplan, Prafulla Dhariwal, Arvind Neelakantan, Pranav Shyam, Girish Sastry, Amanda Askell, Sandhini Agarwal, Ariel Herbert-Voss, Gretchen Krueger, Tom Henighan, Rewon Child, Aditya Ramesh, Daniel~M. Ziegler, Jeffrey Wu, Clemens Winter, Christopher Hesse, Mark Chen, Eric Sigler, Mateusz Litwin, Scott Gray, Benjamin Chess, Jack Clark, Christopher Berner, Sam McCandlish, Alec Radford, Ilya Sutskever, and Dario Amodei. 2020.
\newblock \href {https://arxiv.org/abs/2005.14165} {Language models are few-shot learners}.
\newblock \emph{Preprint}, arXiv:2005.14165.

\bibitem[{Chen et~al.(2021)Chen, Watanabe, Villalba, Żelasko, and Dehak}]{nartrans}
Nanxin Chen, Shinji Watanabe, Jesús Villalba, Piotr Żelasko, and Najim Dehak. 2021.
\newblock \href {https://doi.org/10.1109/LSP.2020.3044547} {Non-autoregressive transformer for speech recognition}.
\newblock \emph{IEEE Signal Processing Letters}, 28:121--125.

\bibitem[{Chi et~al.(2021)Chi, Salazar, and Kirchhoff}]{chi-etal-2021-align}
Ethan~A. Chi, Julian Salazar, and Katrin Kirchhoff. 2021.
\newblock \href {https://doi.org/10.18653/v1/2021.naacl-main.154} {Align-refine: Non-autoregressive speech recognition via iterative realignment}.
\newblock In \emph{Proceedings of the 2021 Conference of the North American Chapter of the Association for Computational Linguistics: Human Language Technologies}, pages 1920--1927, Online. Association for Computational Linguistics.

\bibitem[{Chiu et~al.(2018)Chiu, Sainath, Wu, Prabhavalkar, Nguyen, Chen, Kannan, Weiss, Rao, Gonina, Jaitly, Li, Chorowski, and Bacchiani}]{8462105}
Chung-Cheng Chiu, Tara~N. Sainath, Yonghui Wu, Rohit Prabhavalkar, Patrick Nguyen, Zhifeng Chen, Anjuli Kannan, Ron~J. Weiss, Kanishka Rao, Ekaterina Gonina, Navdeep Jaitly, Bo~Li, Jan Chorowski, and Michiel Bacchiani. 2018.
\newblock \href {https://doi.org/10.1109/ICASSP.2018.8462105} {State-of-the-art speech recognition with sequence-to-sequence models}.
\newblock In \emph{2018 IEEE International Conference on Acoustics, Speech and Signal Processing (ICASSP)}, pages 4774--4778.

\bibitem[{Devlin et~al.(2019)Devlin, Chang, Lee, and Toutanova}]{devlin-etal-2019-bert}
Jacob Devlin, Ming-Wei Chang, Kenton Lee, and Kristina Toutanova. 2019.
\newblock \href {https://doi.org/10.18653/v1/N19-1423} {{BERT}: Pre-training of deep bidirectional transformers for language understanding}.
\newblock In \emph{Proceedings of the 2019 Conference of the North {A}merican Chapter of the Association for Computational Linguistics: Human Language Technologies, Volume 1 (Long and Short Papers)}, pages 4171--4186, Minneapolis, Minnesota. Association for Computational Linguistics.

\bibitem[{Dong et~al.(2018)Dong, Xu, and Xu}]{8462506}
Linhao Dong, Shuang Xu, and Bo~Xu. 2018.
\newblock \href {https://doi.org/10.1109/ICASSP.2018.8462506} {Speech-transformer: A no-recurrence sequence-to-sequence model for speech recognition}.
\newblock In \emph{2018 IEEE International Conference on Acoustics, Speech and Signal Processing (ICASSP)}, pages 5884--5888.

\bibitem[{Futami et~al.(2022)Futami, Inaguma, Mimura, Sakai, and Kawahara}]{futami2022distillingknowledgebertctcbased}
Hayato Futami, Hirofumi Inaguma, Masato Mimura, Shinsuke Sakai, and Tatsuya Kawahara. 2022.
\newblock \href {https://arxiv.org/abs/2209.02030} {Distilling the knowledge of bert for ctc-based asr}.
\newblock \emph{Preprint}, arXiv:2209.02030.

\bibitem[{Graves et~al.(2006)Graves, Fern\'{a}ndez, Gomez, and Schmidhuber}]{ctc-graves}
Alex Graves, Santiago Fern\'{a}ndez, Faustino Gomez, and J\"{u}rgen Schmidhuber. 2006.
\newblock \href {https://doi.org/10.1145/1143844.1143891} {Connectionist temporal classification: labelling unsegmented sequence data with recurrent neural networks}.
\newblock In \emph{Proceedings of the 23rd International Conference on Machine Learning}, ICML '06, page 369–376, New York, NY, USA. Association for Computing Machinery.

\bibitem[{Graves and Jaitly(2014)}]{pmlr-v32-graves14}
Alex Graves and Navdeep Jaitly. 2014.
\newblock \href {https://proceedings.mlr.press/v32/graves14.html} {Towards end-to-end speech recognition with recurrent neural networks}.
\newblock In \emph{Proceedings of the 31st International Conference on Machine Learning}, volume~32 of \emph{Proceedings of Machine Learning Research}, pages 1764--1772, Bejing, China. PMLR.

\bibitem[{Gulati et~al.(2020)Gulati, Qin, Chiu, Parmar, Zhang, Yu, Han, Wang, Zhang, Wu, and Pang}]{gulati2020conformer}
Anmol Gulati, James Qin, Chung-Cheng Chiu, Niki Parmar, Yu~Zhang, Jiahui Yu, Wei Han, Shibo Wang, Zhengdong Zhang, Yonghui Wu, and Ruoming Pang. 2020.
\newblock \href {https://doi.org/10.21437/Interspeech.2020-3015} {Conformer: Convolution-augmented transformer for speech recognition}.
\newblock In \emph{Proc. Interspeech 2020}, pages 5036--5040.

\bibitem[{He et~al.(2016)He, Zhang, Ren, and Sun}]{deepres}
Kaiming He, Xiangyu Zhang, Shaoqing Ren, and Jian Sun. 2016.
\newblock \href {https://doi.org/10.1109/CVPR.2016.90} {Deep residual learning for image recognition}.
\newblock In \emph{2016 IEEE Conference on Computer Vision and Pattern Recognition (CVPR)}, pages 770--778.

\bibitem[{Higuchi et~al.(2021)Higuchi, Chen, Fujita, Inaguma, Komatsu, Lee, Nozaki, Wang, and Watanabe}]{comparative}
Yosuke Higuchi, Nanxin Chen, Yuya Fujita, Hirofumi Inaguma, Tatsuya Komatsu, Jaesong Lee, Jumon Nozaki, Tianzi Wang, and Shinji Watanabe. 2021.
\newblock \href {https://doi.org/10.1109/ASRU51503.2021.9688157} {A comparative study on non-autoregressive modelings for speech-to-text generation}.
\newblock In \emph{2021 IEEE Automatic Speech Recognition and Understanding Workshop (ASRU)}, pages 47--54.

\bibitem[{Higuchi et~al.(2022)Higuchi, Karube, Ogawa, and Kobayashi}]{hierarchical-ctc}
Yosuke Higuchi, Keita Karube, Tetsuji Ogawa, and Tetsunori Kobayashi. 2022.
\newblock \href {https://doi.org/10.1109/ICASSP43922.2022.9746580} {Hierarchical conditional end-to-end asr with ctc and multi-granular subword units}.
\newblock In \emph{ICASSP 2022 - 2022 IEEE International Conference on Acoustics, Speech and Signal Processing (ICASSP)}, pages 7797--7801.

\bibitem[{Higuchi et~al.(2020)Higuchi, Watanabe, Chen, Ogawa, and Kobayashi}]{higuchi2020mask}
Yosuke Higuchi, Shinji Watanabe, Nanxin Chen, Takaaki Ogawa, and Tetsuji Kobayashi. 2020.
\newblock \href {https://doi.org/10.21437/Interspeech.2020-2404} {Mask ctc: Non-autoregressive end-to-end asr with ctc and mask predict}.
\newblock In \emph{Proc. Interspeech 2020}, pages 3655--3659.

\bibitem[{Karita et~al.(2019)Karita, Soplin, Watanabe, Delcroix, Ogawa, and Nakatani}]{karita2019improving}
Shigeki Karita, Nelson Enrique~Yalta Soplin, Shinji Watanabe, Marc Delcroix, Akinori Ogawa, and Tomohiro Nakatani. 2019.
\newblock \href {https://doi.org/10.21437/Interspeech.2019-1938} {Improving transformer-based end-to-end speech recognition with connectionist temporal classification and language model integration}.
\newblock In \emph{Proc. Interspeech 2019}, pages 1408--1412.

\bibitem[{Kim et~al.(2022)Kim, Gholami, Shaw, Lee, Mangalam, Malik, Mahoney, and Keutzer}]{squeeze}
Sehoon Kim, Amir Gholami, Albert Shaw, Nicholas Lee, Karttikeya Mangalam, Jitendra Malik, Michael~W. Mahoney, and Kurt Keutzer. 2022.
\newblock Squeezeformer: an efficient transformer for automatic speech recognition.
\newblock In \emph{Proceedings of the 36th International Conference on Neural Information Processing Systems}, NIPS '22, Red Hook, NY, USA. Curran Associates Inc.

\bibitem[{Kriman et~al.(2020)Kriman, Beliaev, Ginsburg, Huang, Kuchaiev, Lavrukhin, Leary, Li, and Zhang}]{9053889}
Samuel Kriman, Stanislav Beliaev, Boris Ginsburg, Jocelyn Huang, Oleksii Kuchaiev, Vitaly Lavrukhin, Ryan Leary, Jason Li, and Yang Zhang. 2020.
\newblock \href {https://doi.org/10.1109/ICASSP40776.2020.9053889} {Quartznet: Deep automatic speech recognition with 1d time-channel separable convolutions}.
\newblock In \emph{ICASSP 2020 - 2020 IEEE International Conference on Acoustics, Speech and Signal Processing (ICASSP)}, pages 6124--6128.

\bibitem[{Lee and Watanabe(2021)}]{ctc-inter}
Jaesong Lee and Shinji Watanabe. 2021.
\newblock \href {https://doi.org/10.1109/ICASSP39728.2021.9414594} {Intermediate loss regularization for ctc-based speech recognition}.
\newblock In \emph{ICASSP 2021 - 2021 IEEE International Conference on Acoustics, Speech and Signal Processing (ICASSP)}, pages 6224--6228.

\bibitem[{Nozaki and Komatsu(2021)}]{nozaki2021relaxing}
Jun Nozaki and Takashi Komatsu. 2021.
\newblock \href {https://doi.org/10.21437/Interspeech.2021-911} {Relaxing the conditional independence assumption of ctc-based asr by conditioning on intermediate predictions}.
\newblock In \emph{Proc. Interspeech 2021}, pages 3735--3739.

\bibitem[{{OpenAI et al.}(2024)}]{openai2024gpt4technicalreport}
{OpenAI et al.} 2024.
\newblock \href {https://arxiv.org/abs/2303.08774} {Gpt-4 technical report}.
\newblock \emph{Preprint}, arXiv:2303.08774.

\bibitem[{Panayotov et~al.(2015)Panayotov, Chen, Povey, and Khudanpur}]{7178964}
Vassil Panayotov, Guoguo Chen, Daniel Povey, and Sanjeev Khudanpur. 2015.
\newblock \href {https://doi.org/10.1109/ICASSP.2015.7178964} {Librispeech: An asr corpus based on public domain audio books}.
\newblock In \emph{2015 IEEE International Conference on Acoustics, Speech and Signal Processing (ICASSP)}, pages 5206--5210.

\bibitem[{Paul and Baker(1992)}]{paul-baker-1992-design}
Douglas~B. Paul and Janet~M. Baker. 1992.
\newblock \href {https://aclanthology.org/H92-1073/} {The design for the {W}all {S}treet {J}ournal-based {CSR} corpus}.
\newblock In \emph{Speech and Natural Language: Proceedings of a Workshop Held at Harriman, New York, {F}ebruary 23-26, 1992}.

\bibitem[{Radford et~al.(2022)Radford, Kim, Xu, Brockman, McLeavey, and Sutskever}]{radford2022robustspeechrecognitionlargescale}
Alec Radford, Jong~Wook Kim, Tao Xu, Greg Brockman, Christine McLeavey, and Ilya Sutskever. 2022.
\newblock \href {https://arxiv.org/abs/2212.04356} {Robust speech recognition via large-scale weak supervision}.
\newblock \emph{Preprint}, arXiv:2212.04356.

\bibitem[{Rekesh et~al.(2023)Rekesh, Koluguri, Kriman, Majumdar, Noroozi, Huang, Hrinchuk, Puvvada, Kumar, Balam, and Ginsburg}]{fastconfo}
Dima Rekesh, Nithin~Rao Koluguri, Samuel Kriman, Somshubra Majumdar, Vahid Noroozi, He~Huang, Oleksii Hrinchuk, Krishna Puvvada, Ankur Kumar, Jagadeesh Balam, and Boris Ginsburg. 2023.
\newblock \href {https://doi.org/10.1109/ASRU57964.2023.10389701} {Fast conformer with linearly scalable attention for efficient speech recognition}.
\newblock In \emph{2023 IEEE Automatic Speech Recognition and Understanding Workshop (ASRU)}, pages 1--8.

\bibitem[{Rousseau et~al.(2012)Rousseau, Del{\'e}glise, and Est{\`e}ve}]{rousseau-etal-2012-ted}
Anthony Rousseau, Paul Del{\'e}glise, and Yannick Est{\`e}ve. 2012.
\newblock \href {https://aclanthology.org/L12-1405/} {{TED}-{LIUM}: an automatic speech recognition dedicated corpus}.
\newblock In \emph{Proceedings of the Eighth International Conference on Language Resources and Evaluation ({LREC}`12)}, pages 125--129, Istanbul, Turkey. European Language Resources Association (ELRA).

\bibitem[{Touvron et~al.(2023)Touvron, Lavril, Izacard, Martinet, Lachaux, Lacroix, Rozière, Goyal, Hambro, Azhar, Rodriguez, Joulin, Grave, and Lample}]{touvron2023llamaopenefficientfoundation}
Hugo Touvron, Thibaut Lavril, Gautier Izacard, Xavier Martinet, Marie-Anne Lachaux, Timothée Lacroix, Baptiste Rozière, Naman Goyal, Eric Hambro, Faisal Azhar, Aurelien Rodriguez, Armand Joulin, Edouard Grave, and Guillaume Lample. 2023.
\newblock \href {https://arxiv.org/abs/2302.13971} {Llama: Open and efficient foundation language models}.
\newblock \emph{Preprint}, arXiv:2302.13971.

\bibitem[{Vaswani et~al.(2017)Vaswani, Shazeer, Parmar, Uszkoreit, Jones, Gomez, Kaiser, and Polosukhin}]{allyouneed}
Ashish Vaswani, Noam Shazeer, Niki Parmar, Jakob Uszkoreit, Llion Jones, Aidan~N Gomez, \L~ukasz Kaiser, and Illia Polosukhin. 2017.
\newblock Attention is all you need.
\newblock In \emph{Advances in Neural Information Processing Systems}, volume~30. Curran Associates, Inc.

\bibitem[{Wolf et~al.(2020)Wolf, Debut, Sanh, Chaumond, Delangue, Moi, Cistac, Rault, Louf, Funtowicz, Davison, Shleifer, von Platen, Ma, Jernite, Plu, Xu, Scao, Gugger, Drame, Lhoest, and Rush}]{wolf2020huggingfacestransformersstateoftheartnatural}
Thomas Wolf, Lysandre Debut, Victor Sanh, Julien Chaumond, Clement Delangue, Anthony Moi, Pierric Cistac, Tim Rault, Rémi Louf, Morgan Funtowicz, Joe Davison, Sam Shleifer, Patrick von Platen, Clara Ma, Yacine Jernite, Julien Plu, Canwen Xu, Teven~Le Scao, Sylvain Gugger, Mariama Drame, Quentin Lhoest, and Alexander~M. Rush. 2020.
\newblock \href {https://arxiv.org/abs/1910.03771} {Huggingface's transformers: State-of-the-art natural language processing}.
\newblock \emph{Preprint}, arXiv:1910.03771.

\bibitem[{Yang et~al.(2024)Yang, Ma, Gao, Zhang, and Chen}]{ctc-llm}
Guanrou Yang, Ziyang Ma, Zhifu Gao, Shiliang Zhang, and Xie Chen. 2024.
\newblock \href {https://doi.org/10.1109/SLT61566.2024.10832154} {Ctc-assisted llm-based contextual asr}.
\newblock In \emph{2024 IEEE Spoken Language Technology Workshop (SLT)}, pages 126--131.

\bibitem[{Yang et~al.(2023)Yang, Li, and Du}]{ctc-gated}
Yuting Yang, Yuke Li, and Binbin Du. 2023.
\newblock \href {https://doi.org/10.1109/ICASSP49357.2023.10094820} {Improving ctc-based asr models with gated interlayer collaboration}.
\newblock In \emph{ICASSP 2023 - 2023 IEEE International Conference on Acoustics, Speech and Signal Processing (ICASSP)}, pages 1--5.

\end{thebibliography}
\end{raggedright}

\end{document}